\documentclass[letterpaper, 10 pt, journal, twoside]{IEEEtran}
\usepackage{epsfig} 
\usepackage{mathptmx} 
\usepackage{times} 
\usepackage{amsmath} 
\usepackage{amssymb}  
\usepackage{xcolor}
\usepackage{subcaption}
\usepackage[font=small]{caption}
\usepackage{float}
\usepackage{multirow}
\usepackage[flushleft]{threeparttable}
\usepackage{pifont}
\let\labelindent\relax
\usepackage{enumitem}
\usepackage{hyperref}
\usepackage{balance}

\hypersetup{
    colorlinks=true,
    linkcolor=black,
    filecolor=magenta,      
    urlcolor=magenta,
    }

\newcommand{\formattedparagraph}[1]{\noindent \textbf{#1}}

\newcommand{\ie}{{\it{i.e.}}}
\newcommand{\etc}{{\it{etc}}}
\newcommand{\etal}{\textit{et al.}}

\pdfoutput=1

\ifCLASSINFOpdf
\else
\fi
\begin{document}
%
\title{A Real-Time Online Learning Framework for Joint 3D Reconstruction and Semantic Segmentation of Indoor Scenes}
%
%
%

\author{Davide Menini, Suryansh Kumar$^{\dagger}$, Martin R. Oswald, Erik Sandstr\"om, Cristian Sminchisescu, Luc Van Gool 

\thanks{*This work was funded by Focused Research Award from Google.
}%
\thanks{Davide Menini, Suryansh Kumar, Erik Sandstr\"om are with ETH Z\"urich.
}
\thanks{Luc Van Gool is with CVL, ETH Z\"urich and PSI, KU Leuven, Belgium.
}
\thanks{Martin R. Oswald is with CVG Group at ETH Z\"urich.
}
\thanks{Cristian Sminchisescu is with Google Research Z\"urich.
}

\thanks{$\dagger$ Corresponding Author: Suryansh Kumar (k.sur46@gmail.com).
}%
}
%
%

\markboth{IEEE Robotics and Automation Letters. Preprint Version. December, 2021}
{Menini \MakeLowercase{\textit{et al.}}: A Real-Time Online Learning Framework for Joint 3D Reconstruction and Semantic Segmentation of Indoor Scenes} 

%



\maketitle


\begin{abstract}
This paper presents a real-time online vision framework to jointly recover an indoor scene's 3D structure and semantic label. Given noisy depth maps, a camera trajectory, and 2D semantic labels at train time, the proposed deep neural network based approach learns to fuse the depth over frames with suitable semantic labels in the scene space. Our approach exploits the joint volumetric representation of the depth and semantics in the scene feature space to solve this task. For a compelling online fusion of the semantic labels and geometry in real-time, we introduce an efficient vortex pooling block while dropping the use of routing network in online depth fusion to preserve high-frequency surface details. We show that the context information provided by the semantics of the scene helps the depth fusion network learn noise-resistant features. Not only that, it helps overcome the shortcomings of the current online depth fusion method in dealing with thin object structures, thickening artifacts, and false surfaces. Experimental evaluation on the Replica dataset shows that our approach can perform depth fusion at 37 and 10 frames per second with an average reconstruction F-score of 88\% and 91\%, respectively, depending on the depth map resolution. Moreover, our model shows an average IoU score of 0.515 on the ScanNet 3D semantic benchmark leaderboard. {Code and example dataset information is available at {\normalfont\url{https://github.com/suryanshkumar/online-joint-depthfusion-and-semantic}}}.
\end{abstract}


\smallskip

%
\IEEEpeerreviewmaketitle

\section{Introduction}
%
%
%
%
\IEEEPARstart{F}{or} robots to infer 3D geometry of an indoor scene, RGB-D based 3D reconstruction has lately become a popular choice  \cite{newcombe2011kinectfusion} \cite{dai2017bundlefusion} \cite{zollhofer2018state} \cite{Weder2020RoutedFusionLR}. While the RGB-D methods provide reliable 3D geometry of the scene, they are limited to reveal only its structural information. Besides, 3D semantic segmentation methods can label each 3D point with corresponding object categories in the scene space \cite{hou20193d} \cite{tatarchenko2018tangent}, thereby providing a piece of higher-level information about the scene.

For an autonomous robot, it is desirable that its vision and learning module should be capable enough to understand and infer both semantic and scene geometry. Further, the algorithm performing such tasks must be real-time, so that it can be helpful for other important robotic perception and decision making tasks such as scene understanding \cite{hu2021bidirectional}, obstacle detection \cite{kumar2014markov} \cite{mittal2014small}, frontier detection \cite{yamauchi1997frontier} \cite{upadhyay2014crf} \cite{kumar2012bayes}, visual servoing \cite{hutchinson1996tutorial}, online motion planning \cite{han2019fiesta}, \etc.

\begin{figure*}
\centering
\begin{subfigure}{0.32\textwidth}
    \includegraphics[width=1.0\linewidth, height=0.7\linewidth]{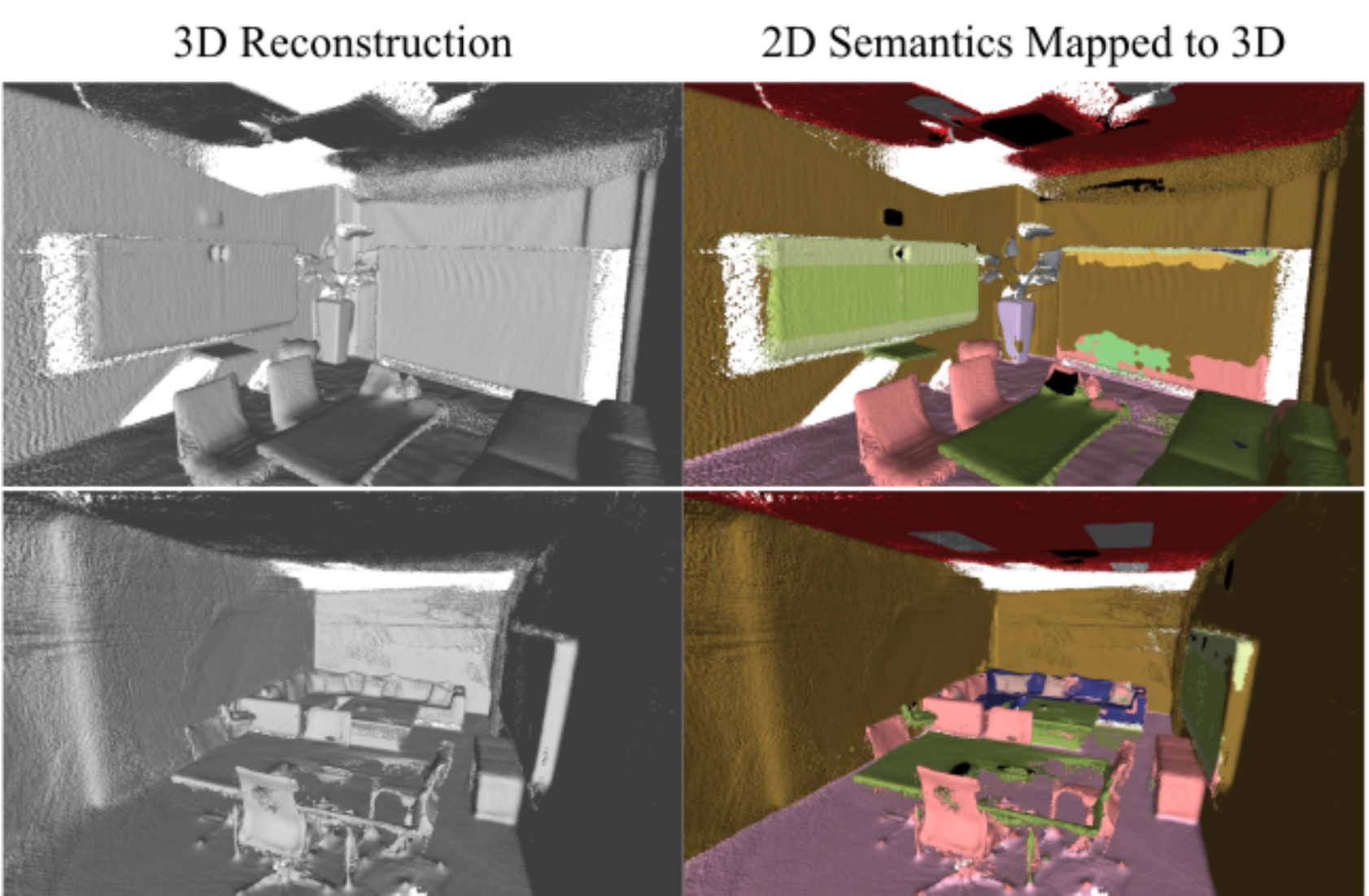}
    \caption{Online depth fusion + 2D semantic projection.} 
    \label{fig:init_figure.a}
\end{subfigure}%
\hspace{0.1cm}
\begin{subfigure}{0.32\textwidth}
   \includegraphics[width=1.0\linewidth, height=0.7\linewidth]{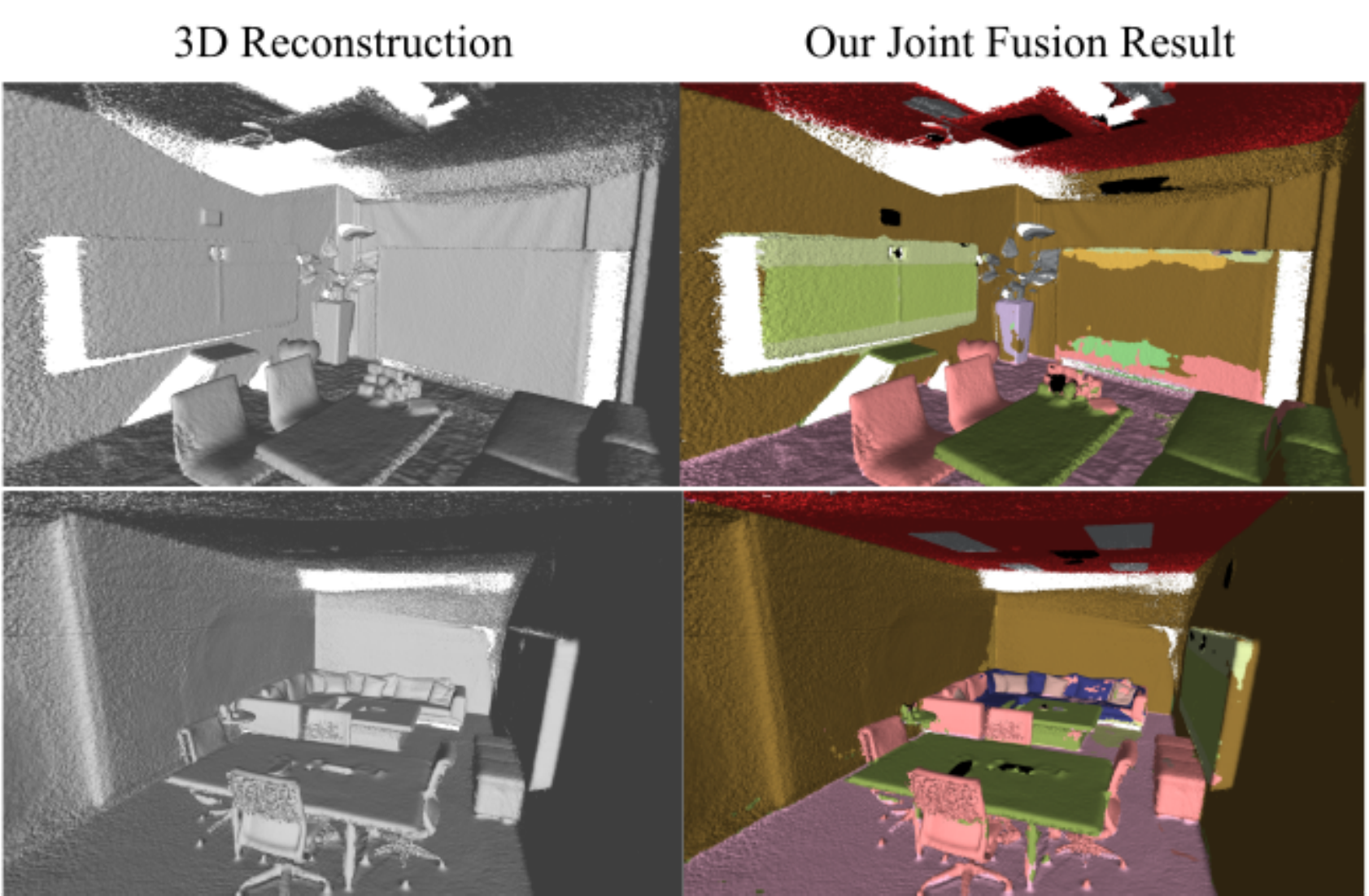}
    \caption{Our Joint framework} 
    \label{fig:init_figure.b}
  \end{subfigure}%
\hspace{0.1cm}
\begin{subfigure}{0.32\textwidth}
   \includegraphics[width=0.92\linewidth, height=0.7\linewidth]{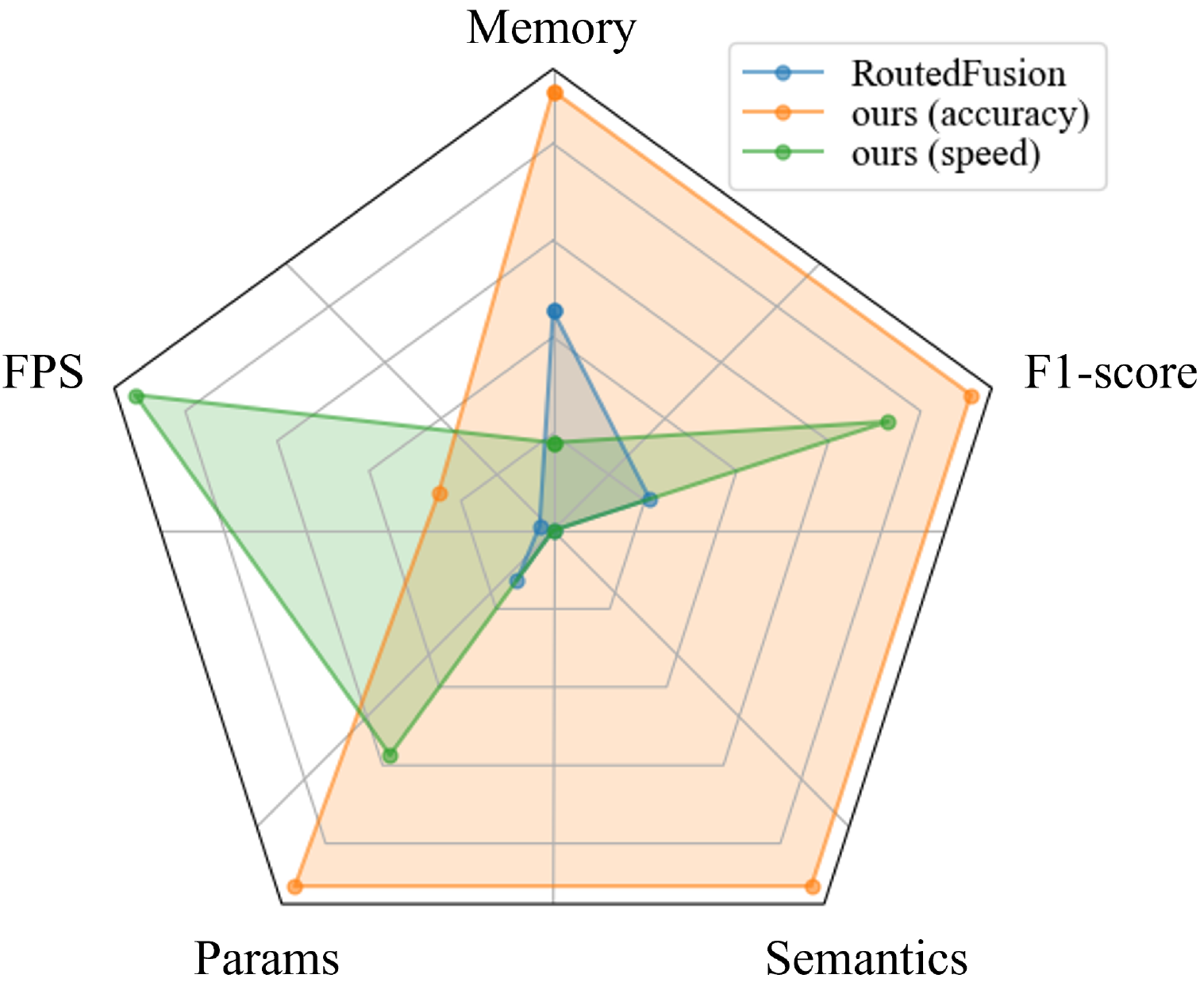}
    \caption{Performance over different metric.} 
    \label{fig:init_figure.c}
  \end{subfigure}%
\smallskip
\caption{(a) \textit{Left:} 3D reconstruction result using recent online depth map fusion method \cite{Weder2020RoutedFusionLR} \textit{Right:} Results via back-projection of the 2D semantic labels to the 3D reconstruction of the scene obtained using \cite{Weder2020RoutedFusionLR}. (b) Better results using our proposed joint framework. (c) RADAR plot showing the visual performance comparison of our method with \cite{Weder2020RoutedFusionLR} over different metrics. \textit{Orange}: Our complex network model gives better reconstruction accuracy (F1-score) with more memory footprint and slightly less FPS. \textit{Green}: Our memory efficient network provides much better depth fusion rate (FPS) and accuracy compared to  \cite{Weder2020RoutedFusionLR} with less model parameters.}
\label{fig:init_figure}
\end{figure*}

Ideally, we want to have an online real-time vision system for a robot \ie, it can process the relevant information on the fly. Unfortunately, the tasks of online 3D reconstruction and 3D semantic labeling are traditionally solved in isolation \cite{ curless1996volumetric} \cite{Weder2020RoutedFusionLR} \cite{ cavallari2016semanticfusion}. For instance, Oleynikova \etal ~\cite{oleynikova2017voxblox} proposed a geometric online RGB-D fusion approach, while  Weder \etal  ~\cite{Weder2020RoutedFusionLR} recently proposed a deep learning-based online framework for depth map fusion demonstrating state-of-the-art results. Further, several learning-based methods have recently emerged that take advantage of 3D structure information to improve 3D semantic label prediction \cite{dai20183dmv} \cite{hu-2021-bidirectional} \cite{choy20194d} \cite{kundu2020virtual}. Yet,  these learning-based 3D semantic labeling methods generally operate offline, hence, unsuitable for robots which requires an online prediction module. Moreover, these methods leverage known geometry to predict semantic labels, and therefore, if the predicted 3D geometry is misleading, the semantic labeling scheme is likely to fail.

We know from the early works in a similar direction that joint inference of depth and semantic labels help the model to take a more informed decision, which is not possible if handled separately \cite{hane2013joint} \cite{sengupta2013urban} \cite{kundu2014joint}. Nevertheless, these methods are offline, restricting their broader adoption for robot vision applications. Instead, this paper proposes an online neural network framework that utilizes the scene's 2D semantics and depth maps.  As a result, our proposed network learns meaningful features at train time and provides more informed priors for joint fusion leading to better performance at test time. At test time, the framework takes the raw depth map and predicted 2D semantic labels for joint inference of 3D geometry and its 3D semantic labels (Sec. \ref{sec:experiment}).

Given the camera intrinsic, poses, and RGB-D as input, we choose to represent the indoor scene space using frame-aligned 3D volumetric grid representation (\emph{voxel}). At each camera-aligned voxel, we store the TSDF value \cite{curless1996volumetric} (\emph{geometry}) and its semantic label, coming from a pre-trained network \cite{valada2019self}, as a separate channel. Updates to the volumetric representation are done by iteratively fusing a set of synchronized depth maps and corresponding 2D semantics labels. We train the proposed depth fusion and 2D semantic segmentation network independently in a supervised manner. For better utilization of depth and semantic data, we introduce the vortex pooling blocks \cite{xie2018vortex} in the network design and exploit the predicted depth map confidence value at each pixel for semantic label update in the 3D scene space.

The intuition is, contrary to depth-sensing modality; 2D semantic prior gives per-pixel label measurement and can provide reliable prior irrespective of surface complexity. Further, it is generally not dependent on the range of the scene. So, the idea is to have a neural network that can learn from complementary sources of information for joint inference of scene geometry and semantic label. Thus, we propose a network that leverages semantic and depth knowledge for feature representation. Our representation is specifically helpful in suppressing noise of the depth maps and infers correct geometry from the scene layout provided by the semantic labels.  Now, it is possible to do the fusion of such complementary information in several complex ways \cite{kundu2014joint} \cite{mccormac2017semanticfusion} \cite{zhang2020fusion}. However, considering the robot's processing power and communication capability with the GPU server, this paper also presents three different variations of the fusion network for usage depending on the computational support.

Experiments on the Replica \cite{straub2019replica} and ScanNet  \cite{dai2017scannet} datasets show that our proposed approach not only improves the 3D reconstruction accuracy but also outperform the SOTA online depth fusion method \cite{Weder2020RoutedFusionLR} on several metric (see Fig.\ref{fig:init_figure}). For training our proposed network, we processed Replica and ScanNet non-watertight meshes to make them watertight. In summary, this paper makes the following contributions.

\begin{itemize}[leftmargin=*]
    \item We present an online approach to jointly infer 3D geometry and 3D semantic labels of an indoor scene. For that, we propose an online depth fusion network, which is faster than RoutedFusion \cite{Weder2020RoutedFusionLR} and better manages noise and high-frequency details. Further, for 3D semantic labeling, we improved on training and loss function of AdapNet++ \cite{valada2019self} for performance gain.
    \item Our online framework is real-time and can provide the results at 10 or 37 FPS (frames per second) depending on input resolution and choice of proposed fusion model.
    \item Our approach consistently show better run-time and 3D reconstruction accuracy than the recent online depth fusion methods\cite{oleynikova2017voxblox}\cite{Weder2020RoutedFusionLR}, when tested on Replica dataset \cite{straub2019replica}. Moreover, our method achieve a mean IoU of 0.515 on the ScanNet 3D semantic segmentation test leaderboard \cite{dai2017scannet}, which is better than many offline methods \cite{dai20183dmv}\cite{rethage2018fully}.
\end{itemize}

%

\section{Related Work}
\noindent
Depth fusion and 3D semantic segmentation for indoor and outdoor scenes have been vastly studied in the past. Nevertheless, these problems are generally solved separately \cite{curless1996volumetric} \cite{Weder2020RoutedFusionLR} \cite{izadi2011kinectfusion} \cite{song2017semantic} \cite{graham20183d}. There exist methods to solve this tasks jointly \cite{kundu2014joint} \cite{reddy2015dynamic} \cite{hane2013joint}. Yet, its relevance to the online robot vision application is somewhat limited and not apt for real-time applications. Since the literature on these two independent topics is extensive, we briefly discuss methods of direct relevance to our work.

\formattedparagraph{Online RGB-D Fusion.}
TSDF fusion \cite{curless1996volumetric} is a  seminal work for fusing noisy depth maps in an online way. Numerous extensions of it like KinectFusion \cite{izadi2011kinectfusion} use it to performs tracking and 3D reconstruction in real-time. Later works tried to improve the RGB-D reconstruction for large-scale scenes with efficient memory usage like voxel hashing \cite{niessner2013real}, octrees \cite{steinbrucker2013large} \cite{fuhrmann2011fusion} \cite{marniok2017efficient},  hierarchical hashing \cite{kahler2015hierarchical}, and grouped raycasting \cite{oleynikova2017voxblox}. For a detailed survey on RGBD-fusion, we refer the reader to Zollh\"ofer \etal ~\cite{zollhofer2018state} paper.  Recently, RoutedFusion \cite{Weder2020RoutedFusionLR} introduced an online learning approach to depth map fusion and demonstrated better performance than other similar methods.

\formattedparagraph{Online 3D Semantic Segmentation.} To our knowledge, popular methods on 3D semantic segmentation are tailored for offline settings \cite{song2017semantic} \cite{graham20183d} \cite{dai20183dmv}. These methods assume surface geometry is provided as input which is not apt for online robot vision applications. Recent online method \cite{zobeidi2020dense} uses Gaussian Processes for Metric-Semantic Mapping that involves complex Bayesian inference framework.

\formattedparagraph{Joint Framework.}  
Methods that jointly solve this problem for robotic applications use dense Conditional Random Field (CRF) models \cite{narita2019panopticfusion} with sparse Structure-from-Motion points, and scene 2D label as input in a SLAM framework \cite{kundu2014joint} \cite{reddy2015dynamic}. \cite{cavallari2016semanticfusion} extends KinectFusion \cite{newcombe2011kinectfusion} to improve tracking and perform 3D semantic labeling on the fly. The 3D semantic labeling is done by projecting 2D semantic predictions into 3D and aggregating labels to each voxel via a voting scheme, which requires full histogram volumes to be stored (\textit{e.g.}, with 30 classes, it requires 30 times more memory than ours). Contrary to \cite{cavallari2016semanticfusion}, CNN-SLAM \cite{tateno2017cnn} uses depth prediction network that takes RGB image as input. The 3D semantic segmentation uses a conceptually similar procedure to fuse 2D semantics in 3D as \cite{cavallari2016semanticfusion}. SemanticFusion \cite{mccormac2017semanticfusion} uses ElasticFusion for geometric reconstruction and computes 2D semantic segmentation of RGB-D frames. The updates to the 3D semantic labels are done by a moving average similar to \cite{cavallari2016semanticfusion}.  However, it applies a CRF model to regularize the labels in 3D. \cite{pham2019real} also uses a CRF model to refine 3D labels predicted online. \cite{zhang2020fusion} proposes a global-local tree dynamic data structure enabling semantic segmentation with RGB-D fusion. Han \etal ~\cite{han2020live} proposed joint framework for AR and mobile devices. \cite{rosinol2020kimera} is a complete SLAM framework with scene representation module as well, and mostly composed of non deep learning approaches. Further, their 3D reconstruction module is based on \cite{oleynikova2017voxblox}. On the contrary, our work is a standalone method that is dedicated to reconstructs the scene geometry and predicts 3D semantic label jointly at a better real-time rate for an online learning-based setting.

\section{Background and Proposed Method}
\noindent
Here, we explain our proposed network in detail. For better understanding, we give a brief review of classic TSDF fusion and its recent extension to the online learning framework.

\subsection{Background.}
TSDF fusion \cite{curless1996volumetric} is an incremental method to integrate the depth maps over frames for each location $\mathbf{x} \in \mathbb{R}^3$ into a volume by averaging truncated signed distance functions (TSDF). Given depth map $D_{t} \in \mathbb{R}^{H\times W}$ from known camera pose $P_t \in SE(3)$ with camera intrinsics $K_t \in \mathbb{R}^{3 \times 3}$ for each time steps $t = 1, 2,\dots,T$, the signed distance update $v_t$ and its corresponding weight $w_t$ integrate the depth measurements $D_t$ at time step $t$ for each location into a discretized TSDF volume $\mathbf{V}_t\in \mathbb{R}^{X\times Y\times Z}$ with weight $\mathbf{W}_t\in \mathbb{R}^{X\times Y\times Z}$ (defined over the entire scene) using the following update equations:

\begin{equation}
\label{eq:tsdf_v}
    \mathbf{V}_t(\mathbf{x}) = \frac{\mathbf{W}_{t-1}(\mathbf{x})\cdot \mathbf{V}_{t-1}(\mathbf{x}) + w_t(\mathbf{x})\cdot v_t(\mathbf{x})}{\mathbf{W}_{t-1}(\mathbf{x}) + w_t(\mathbf{x})}
\end{equation}
\begin{equation}
\label{eq:tsdf_w}
    \mathbf{W}_t(\mathbf{x}) = \mathbf{W}_{t-1}(\mathbf{x}) + w_t(\mathbf{x})
\end{equation}

$\mathbf{V}_t$ and $\mathbf{W}_t$ are initialized to zero. Generally, the update functions are truncated before and after the surface to ensure efficient runtimes and robust reconstruction of fine structures. But, truncation distance is scene-dependent; hence, it is not a general approach for handling anisotropic noise and outliers in the depth maps of any given scene. Concretely, it requires cumbersome fine-tuning of the truncation distance parameter to improve the reconstruction quality.

\begin{figure*}[t]
    \centering
    \includegraphics[width=0.90\linewidth]{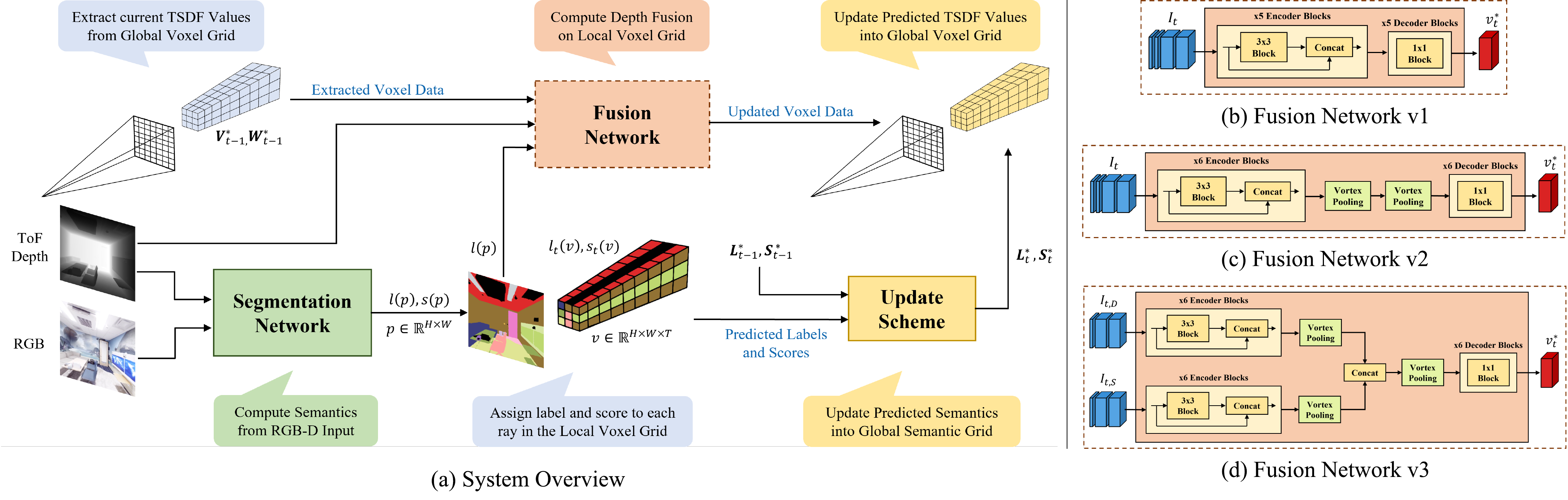}
    \caption{(a) System overview of our method, as thoroughly described in Sec. \ref{ch:method}. (b), (c), (d) are different versions of the Fusion Network.}
    \label{fig:system_overview}
\end{figure*}

\formattedparagraph{Learning based online TSDF Fusion.}
Weder \etal ~\cite{Weder2020RoutedFusionLR} recently proposed RoutedFusion that can overcome the above-mentioned limitations with TSDF fusion. It is a neural network based approach that learn the signed distance update $v_t$ from data.  To achieve the real-time capability, it processes only a sub-volume of the TSDF volume at a time.  Overall, RoutedFusion \cite{Weder2020RoutedFusionLR} is better at handling anisotropic noise that naturally arises from the multi-view setting.  Its pipeline comprises of following 4 steps: \textit{(i) Depth Routing.} The depth routing network takes a noisy depth map $D_t$ and estimate a denoised and outlier corrected depth map $\hat{D}_t$ (routed depth map) for TSDF value extraction. It also provides a per-pixel confidence map $C_t \in \mathbb{R}^{W \times H}$ corresponding to the depth estimate. \textit{(ii) TSDF Extraction.} Given $\hat{D}_t$,  this step extracts a local camera-aligned TSDF values $\mathbf{V}^*_{t-1}\in\mathbb{R}^{H\times W \times T}$ and corresponding weights $\mathbf{W}^*_{t-1}\in\mathbb{R}^{H\times W \times T}$ via trilinear interpolation from the corresponding global voxel grids $\mathbf{V}_{t-1}$ and $\mathbf{W}_{t-1}$. $T$ represents the local discrete extraction window sampled across each ray and centered around each $\hat{D}_t$. \textit{(iii) Depth Fusion.} The depth fusion is performed via a network that takes $\mathbf{V}^*_{t-1}$, $\mathbf{W}^*_{t-1}$ and concatenate it with $\hat{D}_t$ and $C_t$, to construct $\mathbf{I}_{t}$ for subsequent depth fusion:
    \begin{equation*}
        \mathbf{I}_t = [\hat{D}_t \quad C_t \quad  \mathbf{V}^*_{t-1} \quad \mathbf{W}^*_{t-1}] \quad\in\quad\mathbb{R}^{H\times W \times (2T+2)}
    \end{equation*}
    The network consists of an encoding stage to encode local and global information in the viewing frustum, followed by a decoding stage that predicts the TSDF updates $v^*_t\in\mathbb{R}^{H\times W \times T}$ along each ray. \textit{(iv) TSDF Update Integration.} To compute the updated global TSDF  $\mathbf{V}_t$, the predicted local TSDF updates $v_t^*$ are transferred back into the global coordinate frame $v_t$ by inverting the extraction step, \ie, the values are distributed using the same trilinear interpolation weights. $\mathbf{W}_t$ accumulates the splatting weights for each voxel in the scene and is used to filter out extreme outliers in a post-processing step. The integration happens via the update equations Eq. \eqref{eq:tsdf_v} and Eq. \eqref{eq:tsdf_w} only for non-outlier values in $\hat{D}_t$, \ie, values where $C_t > C_\text{th}$. Here, $C_\text{th}$ denotes threshold confidence value.

\subsection{Proposed Approach.}
\label{ch:method}
Our method focuses on developing an online framework that can estimate the indoor scene 3D structure with its proper semantic label in real-time. Fig.(\ref{fig:system_overview}) shows the overall components of our network design. Next, we describe each component separately for better comprehension.

\noindent
\emph{(i) For improved 3D reconstruction}:  
For this, we first propose some modifications to the recent state-of-the-art online depth fusion network \cite{Weder2020RoutedFusionLR}. Our innovations provide significant improvement in the inference time along with better 3D reconstruction accuracy. Next, our proposed \textbf{Fusion Network} uses the predicted 2D semantic label prior, depth-sensor prior, and voxel data of the current frame for efficient depth fusion and voxel grid update. Where, 2D semantic label prior is obtained using AdapNet++ \cite{valada2019self} architecture that takes RGB-D data as input. The use of 2D semantic labels help enhance the 3D reconstruction further.

\noindent
\emph{(ii) For an efficient 3D scene labeling}:
Given a 2D semantic label and its predicted confidence value \cite{valada2019self}, we assign to each ray these values in the updated local voxel grid representation. The \textbf{Update Scheme} takes the current frame labels and their associated confidence score to update the predicted semantics into the global voxel grid representation. 

\smallskip
For the proposed joint framework to be usable across platform with various computing resources, we propose three variations of the \textbf{Fusion Network} (see Figure \ref{fig:system_overview}(b), \ref{fig:system_overview}(c), \ref{fig:system_overview}(d)). Further, we modified the training strategy and loss function of the \textbf{Segmentation Network}, though its network design is inspired by \cite{valada2019self}. Next, we describe the three variations of the proposed fusion network in detail, followed by our update scheme for joint inference of 3D reconstruction and scene labels. The details on training the tailored segmentation network are provided in section \S \ref{sec:implementation}.

\smallskip
\formattedparagraph{(a) Fusion Network v1.}
The goal of this network is to utilize the 2D semantic priors for improving the online 3D reconstruction. The first fusion network architecture -- named as ``\emph{\textbf{Fusion Network v1}}'' (see Fig. \ref{fig:system_overview}(b)) -- is inspired from the recently proposed RoutedFusion \cite{Weder2020RoutedFusionLR}. However, in contrast to RoutedFusion, our fusion network has no depth routing network module and uses the 2D semantic input. The reasons for dropping the depth routing module are \textit{(i)} the confidence map provided by it is observed to have no significant effect on the depth fusion; \textit{(ii)} it removes high-frequency surface details along with the noise, which is critical for thin structures. Our depth fusion network uses the extracted TSDF volumes $\mathbf{V}^*_{t-1}$, corresponding weight $\mathbf{W}^*_{t-1}$, and concatenate it with the noisy depth map ${D}_t$ and the predicted semantic frame $\hat{S}_t$, providing $\mathbf{I}_{t}$ for the subsequent depth fusion \ie,
\begin{equation*}
    \mathbf{I}_t = [D_t \quad \hat{S}_t \quad \mathbf{V}^*_{t-1} \quad \mathbf{W}^*_{t-1}] \quad \in\quad\mathbb{R}^{H\times W \times (2T+2)}
\end{equation*}
The fusion network then predicts the local TSDF update $v^*_t\in\mathbb{R}^{H\times W \times T}$ along each ray. Finally, to compute the updated global TSDF volume $\mathbf{V}_t$, the predicted local TSDF updates $v_t^*$ are transferred back into the global coordinate frame $v_t$ by inverting the extraction step. Overall, by incorporating these modifications in the fusion network, we favorably improved the 3D reconstruction accuracy.

\formattedparagraph{(b) Fusion Network v2.} We inferred that increasing the encoder block learning capacity and aggregating more local and contextual information could encourage the network to make better use of the 2D semantic prior. Accordingly, we increase the number of encoder blocks from 5 to 6 to learn deeper features. Each encoder block is composed of two 3$\times$3 convolutions interleaved with batch normalization, leaky ReLU activation, and drop-out layers with 0.2 probability.  

For better aggregation of local and contextual information, we use vortex pooling block \cite{xie2018vortex}. The block is composed of 5 branches that are eventually concatenated, each with different pooling operations (average pooling with rates $1 \times 1$, $3 \times 3$, $9 \times 9$, $27 \times 27$ and global pooling). The pooling layers are cascaded for computational efficiency: instead of doing three parallel pooling operations with rate 3, 9 and 27, it is more efficient to do three sequential $3 \times 3$ average pooling operations and then apply the convolution on the related branch \cite{xie2018vortex}. In this work, we propose a bottleneck structure to save parameters and TFLOPs. The bottleneck uses a $1\times 1$ convolution to compress the input (channel-wise) before the $3 \times 3$ convolution and then another $1 \times 1$ convolution expands the channels to the desired output dimension. By empirical analysis, we found that two vortex pooling blocks gives better solution. The resulting architecture is named ``\emph{\textbf{Fusion Network v2}}'' (see Fig.\ref{fig:system_overview}(c)).

\formattedparagraph{(c) Fusion Network v3.} A slightly different approach to efficiently utilize semantic labels is to have disentangled encoding heads for depth and semantics. The input of each head is a combination of the extracted TSDF values and weights with either the depth frame or the semantic frame:
\begin{equation*}
    \mathbf{I}_{t,D} = [D_t \quad  \mathbf{V}^*_{t-1} \quad \mathbf{W}^*_{t-1}]
    \quad\quad
    \mathbf{I}_{t,S} = [\hat{S}_t \quad  \mathbf{V}^*_{t-1} \quad \mathbf{W}^*_{t-1}]
\end{equation*}
We refer such a fusion network as ``\emph{\textbf{Fusion Network v3}}'' (see Fig.\ref{fig:system_overview}(d)). It has 2 equal encoder heads, each composed of 6 encoder blocks and a vortex pooling module. The encoded features are fused via concatenation, and then they pass through another vortex pooling block. Finally, the decoder outputs the TSDF update, as in the original architecture.

\formattedparagraph{(d) Update Scheme.} 
It utilizes the semantic information in the global scene space efficiently. Since there is a one-to-one correspondence between the pixels in $\hat{S}_t$ and $D_t$ frame, we can assign a semantic label to each ray in the updated local volume $v_t$ to map the predicted frame label into the semantic volume update $l_t$. Since the overall volume indices remain the same, the local update can be transferred back into the global semantic volume $\mathbf{L}_t$ by updating at the same volume locations (TSDF and weights). To decide whether to insert the predicted semantics into the current global volume, we check the confidence score associated with the label.
Denoting as $\mathbf{L}^*_{t-1}$, $\mathbf{S}^*_{t-1}\in\mathbb{R}^{H\times W \times T}$ the old local label and score volumes, and as $l_t$, $s_t\in\mathbb{R}^{H\times W \times T}$ the respective current predictions, the updates for each voxel are found as follows:
\begin{equation}
\mathbf{S}^*_t(v) = \max (s_t(v),\;\mathbf{S}^*_{t-1}(v))
\end{equation}
\begin{equation}
\mathbf{L}^*_t(v) = 
\begin{cases} 
      l_t(v) & s_t(v) \geq \mathbf{S}^*_{t-1}\\
      \mathbf{L}^*_{t-1}(v) & s_t(v) < \mathbf{S}^*_{t-1} \\
   \end{cases}
\end{equation}
We update semantic labels whose confidence is higher than the one previously stored. The new scores are updated with the maximum values. If the old labels are equal to the current ones, then the scores are updated only if the confidence is higher. We are conscious that this update may result in the plateauing of the scoring volume at relatively high values if several highly confidently and incoherent labels get integrated. However, this method does not require expensive memory usage and manages to retain the semantic history, which in general mitigates the possible integration of outlier predictions in the semantic volume.

\section{Implementation}
\label{sec:implementation}
\noindent
We implemented our framework in PyTorch 1.4.0. with Python 3.7. The network runs on a computing machine with NVIDIA GeForce RTX 2080Ti GPU (10 GB memory). Other requirements are access to RGBD sensors and to a pretrained AdapNet++ framework, as well as watertight meshes to generate the ground truth TSDF volumes. We trained and tested our network on two well-known indoor datasets, \ie, Replica \cite{straub2019replica} and ScanNet \cite{dai2017scannet}.

\formattedparagraph{(a) Replica Dataset.}
It consists of 16 high-quality 3D scenes of apartments, rooms, and offices. We generated three different representations of the indoor scene data using 3D meshes provided by the dataset: (1) semantic meshes made watertight via screened Poisson surface reconstruction \cite{kazhdan2013screened}, (2) regular voxel grids with signed distance function and semantics at each discrete position (3) 512 $\times$ 512 RGB-D and ground truth semantic images accounting for 55 different pose trajectories. While Replica originally features 100 classes, we rearranged them into 30 frequently observed semantic classes to have a more balanced dataset. Also the training, validation, and test splits are created trying to keep the distribution of categories balanced.

\formattedparagraph{(b) ScanNet Dataset.}
It is a standard dataset to test an algorithm's 3D semantic segmentation accuracy for an indoor scene. The train and validation sets contain 2.5M RGB-D images accounting for 1512 scans acquired in 707 different spaces. It is collected using hardware-synchronized RGB and depth cameras of an iPad Air 2 at 30Hz. The test set comprises of 100 scenes for online benchmarking that are not public. The dataset provides pixel-level and voxel-level semantic annotations for 20 object categories. Unbalanced class labels, irregular annotations across boundaries and unlabelled far objects make the 3D semantic segmentation task challenging on this dataset.

We used this dataset to report our 3D semantic segmentation result. We make few meshes watertight for training and compute the SDF voxel grid as ground truth. Note that watertight mesh helps avoid artifacts in SDF representation.

\subsection{Training}
Due to the non-differentiable nature of the voxel grid feature extraction block, we trained the segmentation and fusion networks independently in a supervised setting.

\formattedparagraph{(a) Segmentation Network.} It is trained in two stages: first, the unimodal AdapNet++ models are trained independently on the respective data, \ie, RGB or depth, then pre-trained encoders are used to train the multimodal architecture. We use the SGD optimizer with Nesterov momentum, learning rate $= 5e^{-3}$,  weight decay $= 5e^{-4}$, batch size $= 8$, and a polynomial scheduler that decreases the learning rate after each epoch. The network is trained with 40K $256 \times 256$ images from Replica and 1M $320 \times 240 $ images for ScanNet. 

For training, we used the sum of $\mathcal{L}_{main}$, $\mathcal{L}_{aux1}$, and $ \mathcal{L}_{aux2}$ loss function. $\mathcal{L}_{main}$ denotes the bootstrap cross-entropy loss between the RGB-D input and the ground-truth 2D semantic segmentation image. $\mathcal{L}_{aux1}$, $\mathcal{L}_{aux2}$ is the same bootstrap cross-entropy loss function but computed at a different scale.  These auxiliary losses embed multi-scale information in the final loss function and accelerate learning.

\begin{equation}
    \mathcal{L}_S = \mathcal{L}_{main} + \lambda_1\mathcal{L}_{aux1} + \lambda_2\mathcal{L}_{aux2}
\end{equation}

We set $\lambda_1=0.6$ and $\lambda_2=0.5$. Contrary to the loss proposed in AdapNet++ \cite{valada2019self}, we use bootstrapped cross-entropy loss, which focuses on  difficult predictions by discarding the easiest ones with a threshold. For each image in the batch, the cross-entropy loss $H$ is calculated and sorted. Given that $N$ is the index of last value greater than the loss threshold $H_\text{th}$ and $K\leq N$ is the minimum number of values to be considered, the bootstrapped cross-entropy for an image is computed as: 
\begin{equation}
\mathcal{L} =
\begin{cases} 
       \sum_{j=1}^{N}{H_j} & H_{K}\geq H_\text{th} \\
      \sum_{j=1}^{K}{H_j} & H_{K}< H_\text{th} \\
   \end{cases}
\end{equation}
We set $K = 4096$ and $H_\text{th} = 0.5$ at train time.

\formattedparagraph{(b) Fusion Network.}
We used 6 and 10 scenes from the Replica and ScanNet benchmark to train this network. At train time, we estimate the loss between the ground truth SDF voxel grid and the estimated SDF voxel grid (computed using noisy ToF depth maps and semantic input frames). The batch size is set to 1 due to the incremental nature of the pipeline. Here, every iteration updates many voxels, which are later used to compute the loss. Stability during training is further enforced by accumulating the gradients every 8 steps before the backward pass. We used RMSprop optimizer (weight decay $0.01$, momentum $0.9$, initial learning rate $1e^{-5}$) to minimize the differences between the updated local TSDF volume $\hat{V}_t^*\in \mathbb{R}^{H\times W \times T}$ and the corresponding ground truth volume $V^*\in \mathbb{R}^{H\times W \times T}$. We set extraction window $T=9$. The total loss is a weighted combination of (1) ${l}_1$ loss for preserving the surface details; (2) $l_2$ loss, which minimizes large surface deviations; (3) cosine embedding loss $l_c$ that ensures that the surface is located at the zero-crossing of the signed distance field. Only valid rays are considered in the loss, \ie, rays corresponding to pixels with a valid depth. For each valid ray $i$ at iteration $t$, the fusion loss is defined as:

\begin{equation}
    \mathcal{L}_F = \sum_i{\lambda_1 {l}_1(\hat{V}_{t,i}^*,V_i^*) + \lambda_2 {l}_2(\hat{V}_{t,i}^*,V_i^*) + \lambda_3 {l}_c(\hat{V}_{t,i}^*,V_i^*)}
\end{equation}
where, we set $\lambda_1 = 1$, $\lambda_2 = 10$ and $\lambda_3 = 0.1$ at train time.

\begin{figure}
    \centering
    \includegraphics[width=0.99\linewidth]{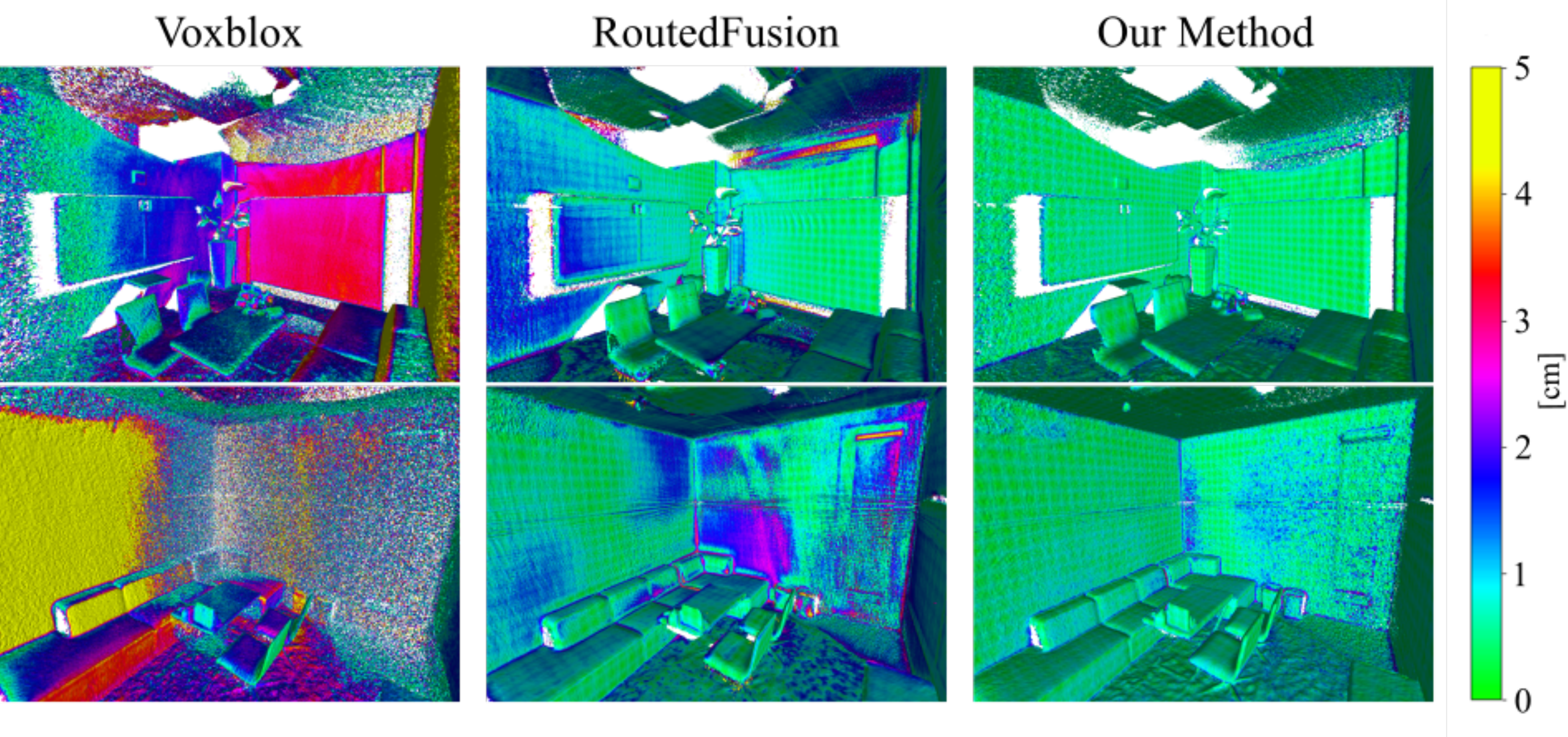}    
    \caption{Qualitative comparison of our method with \cite{oleynikova2017voxblox} and \cite{Weder2020RoutedFusionLR}. This scene is taken from office\_0 of Replica dataset. Color bar show the error, \textit{for e.g.},
    green points have less than $1cm$ error w.r.t GT.}
    \label{fig:prec}
\end{figure}

\subsection{Host-Device Communication}
For improving the runtime on GPU, host-device communication is critical. In RoutedFusion \cite{Weder2020RoutedFusionLR}, the whole database (TSDF weights, TSDF values, \etc) is stored on CPU as dense {\small\texttt{float16}} tensors. Hence, during the extraction and integration steps, the volumes are inefficiently moved between GPU and CPU for computations without considering long delay due to vast data transfer. We tackle this issue by reducing the amount of data transfer between host and device.

\formattedparagraph{Efficient use of computational resources.} 
We only move the extracted volumes to GPU to compute updates. Once the update is done, it is moved back to the host, where the global volumes are located. Computations are performed with {\small\texttt{float32}} precision, but data transfer and storage happens with {\small\texttt{float16}} tensors. Our solution moves only a few MB of data, thereby saving memory and expensive communication without sacrificing the precision. Further, since the database is still stored on the CPU, we can process large scenes without performance loss. An additional improvement comes from memory pinning, enabling faster host-to-device communication when loading the current batch. Indeed the CPU reserves some physical space on its memory (pinned memory) so that the GPU can access it directly. Otherwise, the CPU has to look up the physical address from the virtual address before each CUDA Memcpy operation.

Even with the discussed improvements most of execution time is still spent on data transfer. However, if the users have an high-end on-board GPU, then they can gain extra mileage from our implementation at test time for inference (more than 30 FPS) by storing all volumes on GPU.

\begin{figure*}[t]
    \centering
    \begin{subfigure}{0.30\textwidth}
    \includegraphics[width=\linewidth]{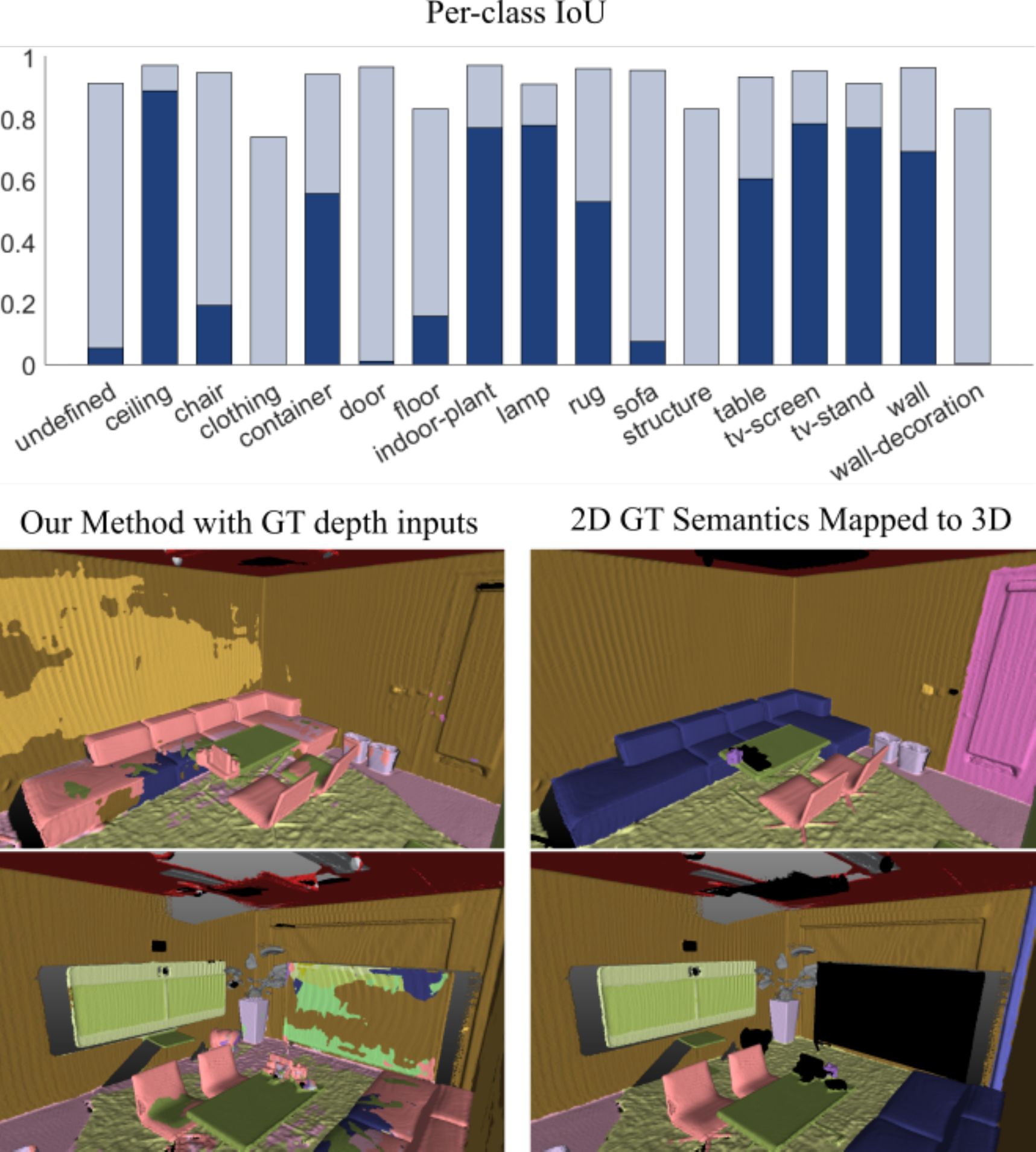}
    \caption{office\_0}
    \end{subfigure}%
    \hspace{0.15cm}
    \begin{subfigure}{0.30\textwidth}
    \includegraphics[width=\linewidth]{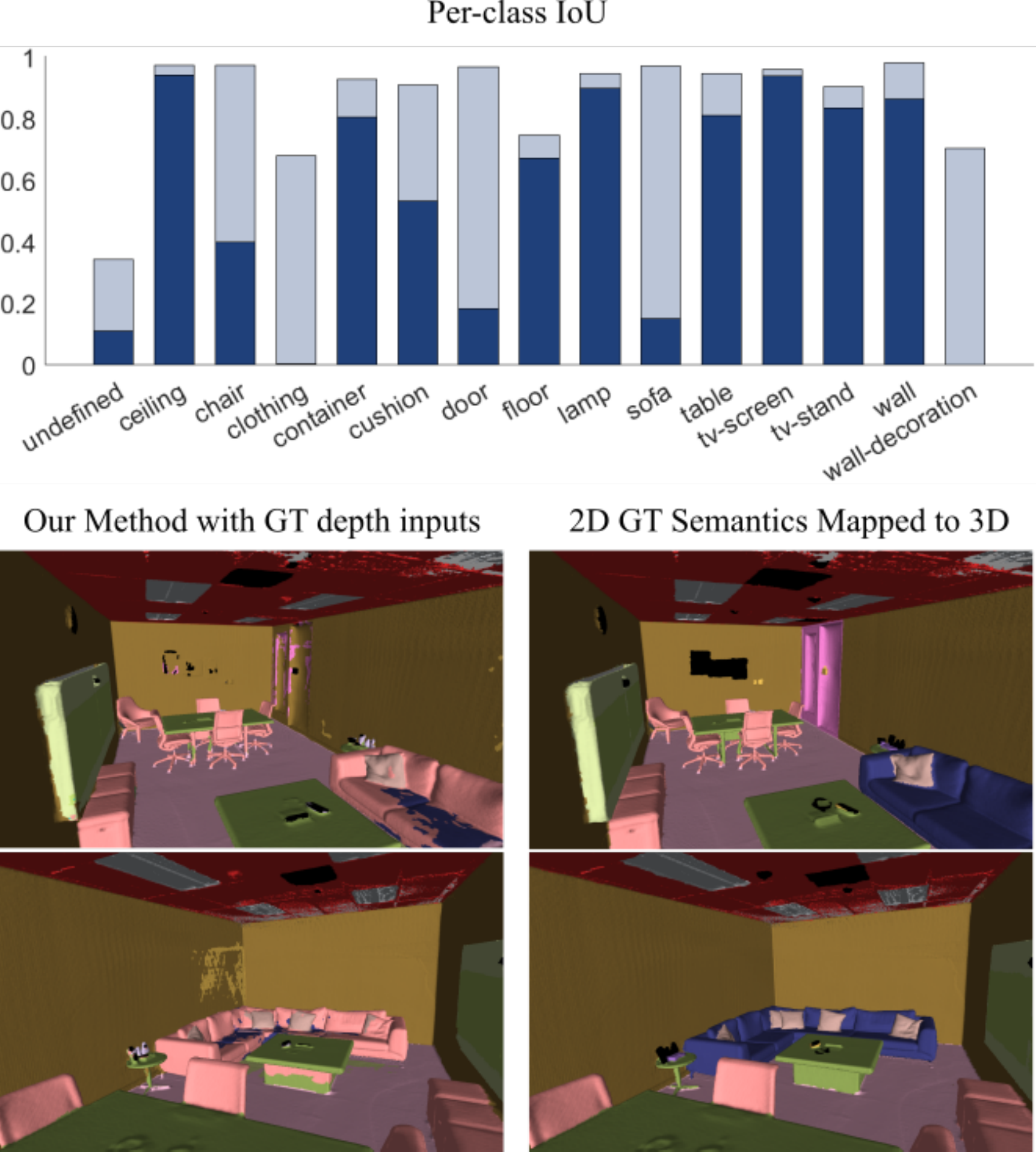}
    \caption{office\_2}
    \end{subfigure}%
    \hspace{0.15cm}
    \begin{subfigure}{0.30\textwidth}
    \includegraphics[width=\linewidth]{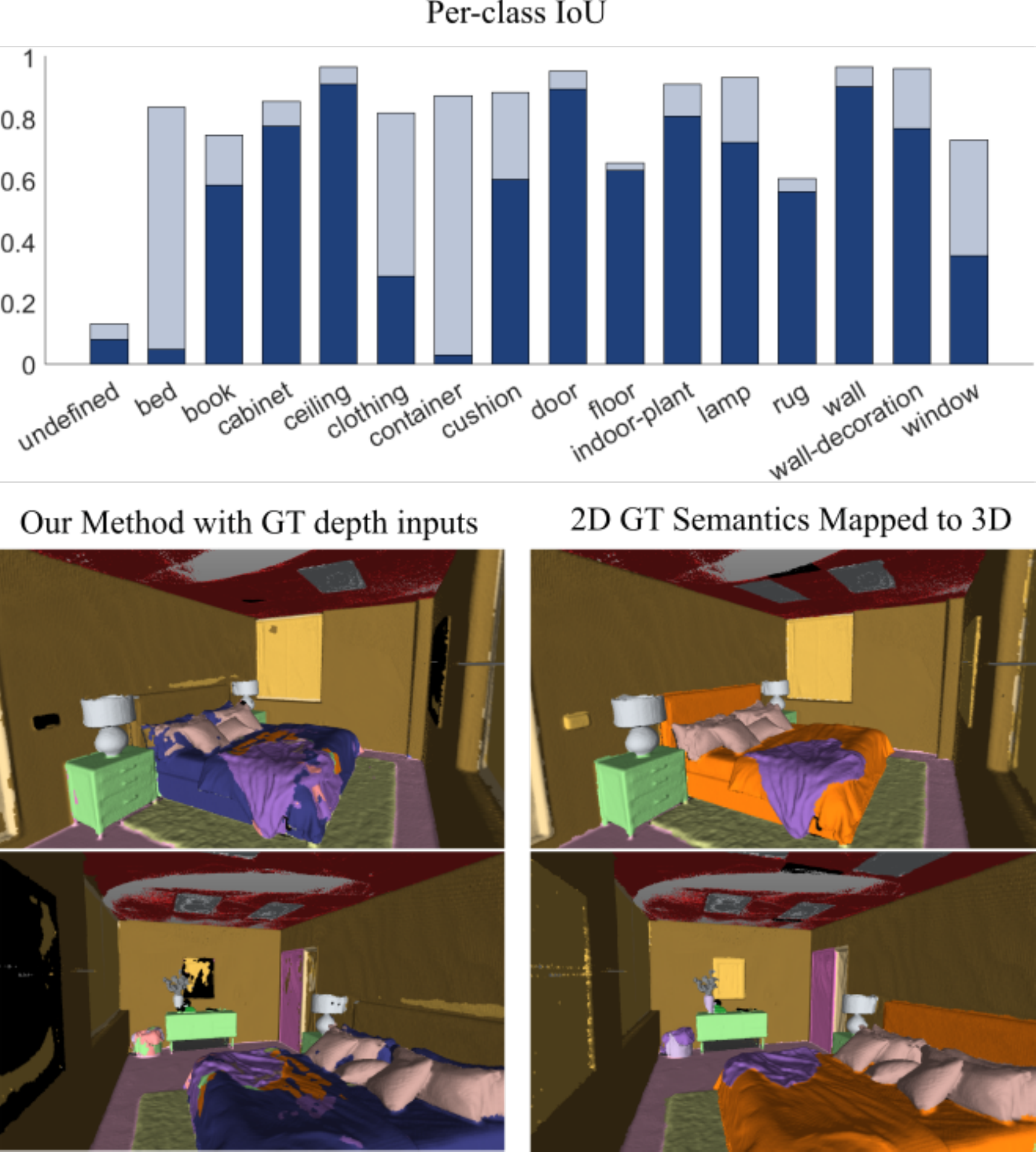}
    \caption{room\_1}
    \end{subfigure}
    \caption{\textbf{Top row}: per class IoU on Replica dataset. Light and dark blue bar shows the synthetically created GT and our prediction results, respectively. For each scene the mean IoU to its upper bound is: (a) 40.29 / 91.34, (b) 54.14 / 86.09, (c) 55.09 / 80.06. \textbf{Bottom row}: Qualitative results for the same on different indoor scenes taken from Replica dataset.}
    \label{fig:seg_comp}
\end{figure*}

\section{Experimental Evaluation and Results}\label{sec:experiment}

\formattedparagraph{Evaluation Metrics.} 
For 3D scene reconstruction, we used F-score metric as in \cite{knapitsch2017tanks}:  $\text{F-score} = 2\cdot\frac{\text{precision}\; \cdot\;  \text{recall}}{\text{precision}+\text{recall}}$. For 3D semantic segmentation evaluation, we compute the per-class Intersection-over-Union (IoU) on the labels associated with the ground truth mesh vertices (for correct geometry) after mapping the estimated and target semantic volumes on them. After computing true positives $TP$, false positives $FP$ and false negatives $FN$ on the confusion matrix, the IoU is calculated as: $\text{IoU} = \frac{TP}{TP+FP+FN}$.

\formattedparagraph{(a) 3D Reconstruction Evaluation.} 
In Table \ref{tab:3d_rec} we compare our 3D reconstruction accuracy with seminal TSDF Fusion \cite{curless1996volumetric}, Voxblox \cite{oleynikova2017voxblox} and RoutedFusion \cite{Weder2020RoutedFusionLR}: our best approach gives 7\% F1-score improvement over the latter (tested on office\_0 scene of Replica dataset). Fig.(\ref{fig:prec}) shows a qualitative comparison with \cite{oleynikova2017voxblox} and \cite{Weder2020RoutedFusionLR}, validating our claim.

\begin{table}[h]
    \centering
    \resizebox{0.7\columnwidth}{!}
    {
    \begin{threeparttable}
    \begin{tabular}{c|c|c|c}
    \hline
        Method & Precision & Recall & F1-score \\
        \hline
        TSDF Fusion \cite{curless1996volumetric} & 73.92 & 69.45 & 71.62 \\
        Voxblox \cite{oleynikova2017voxblox} & 51.39 & 74.42 & 60.80 \\
        RoutedFusion \cite{Weder2020RoutedFusionLR} & 86.95 & 81.66 & 84.22 \\
        \hline
        Ours (speed) &
        \textbf{\textcolor{brown}{96.87}} & \textbf{\textcolor{brown}{83.41}} & \textbf{\textcolor{brown}{89.64}} \\
        Ours (accuracy) & \textbf{98.86} & \textbf{84.54} & \textbf{91.14} \\
    \hline
    \end{tabular}
    \smallskip
    \end{threeparttable}
    }\caption{3D reconstruction result comparison on office\_0.}\label{tab:3d_rec}
\end{table}
\begin{table}[h]
    \centering
    \begin{threeparttable}
    \begin{tabular}{c|c|c|c|c|c}
        \hline
        Model &  Routing & Semantics & Resolution & FPS & mean-F1 \\
        \hline
        RoutedFusion &  \ding{51} & & 256$\times$256 & 1.2 & 82.82\% \\
        \hline
        Fusion Network v1 &  &  & 256$\times$256 & 25.3 & 84.11\% \\
        Fusion Network v1 &  & \ding{51} & 256$\times$256 & 13.8 & 84.53\%\\
        Fusion Network v2 &  & \ding{51} & 256$\times$256 & 11.4 & 89.10\%\\
        Fusion Network v2 & & & 256$\times$256 & 16.2 & 88.32\%\\
        Fusion Network v3 & & \ding{51} & 256$\times$256 & \textbf{10.3} & \textbf{90.35\%}\\
        \hline
        Fusion Network v1 &  &  & 128$\times$128 & 50.5 & 83.71\%\\
        Fusion Network v2 &  &  & 128$\times$128 & \textbf{\textcolor{brown}{37.6}} & \textbf{\textcolor{brown}{88.30\%}}\\
        \hline
    \end{tabular}
    \smallskip
    \caption{Runtime comparison. The best FPS model keeps all data on GPU, avoiding host-device communication. When semantics is enabled, the model uses the predicted semantic frame as additional input to improve the reconstructed geometry.} \label{tab:fusion_runtimes}
    \end{threeparttable}
\end{table}
In Table \ref{tab:3d_rec}, ``Ours (accuracy)'' shows the statistics using Fusion Network v3 and ground truth 2D semantics with $256 \times 256$ input resolution. It runs at 10.3 FPS with a 90.35\% mean F1-score. Whereas ``Ours (speed)'' shows the results of Fusion Network v2 when no 2D semantic is employed. It uses $128 \times 128$ input resolution and runs at 37.6 FPS with a 88.30\% mean F1-score. Runtimes in Table \ref{tab:fusion_runtimes} show that using a smaller input resolution speed up the reconstruction, while providing enough information to the network to keep accuracy high, especially when more complex architectures are used. Disabling semantics further increases speed.

\begin{figure*}[t]
    \centering
    \begin{subfigure}{0.32\textwidth}
    \includegraphics[width=\linewidth]{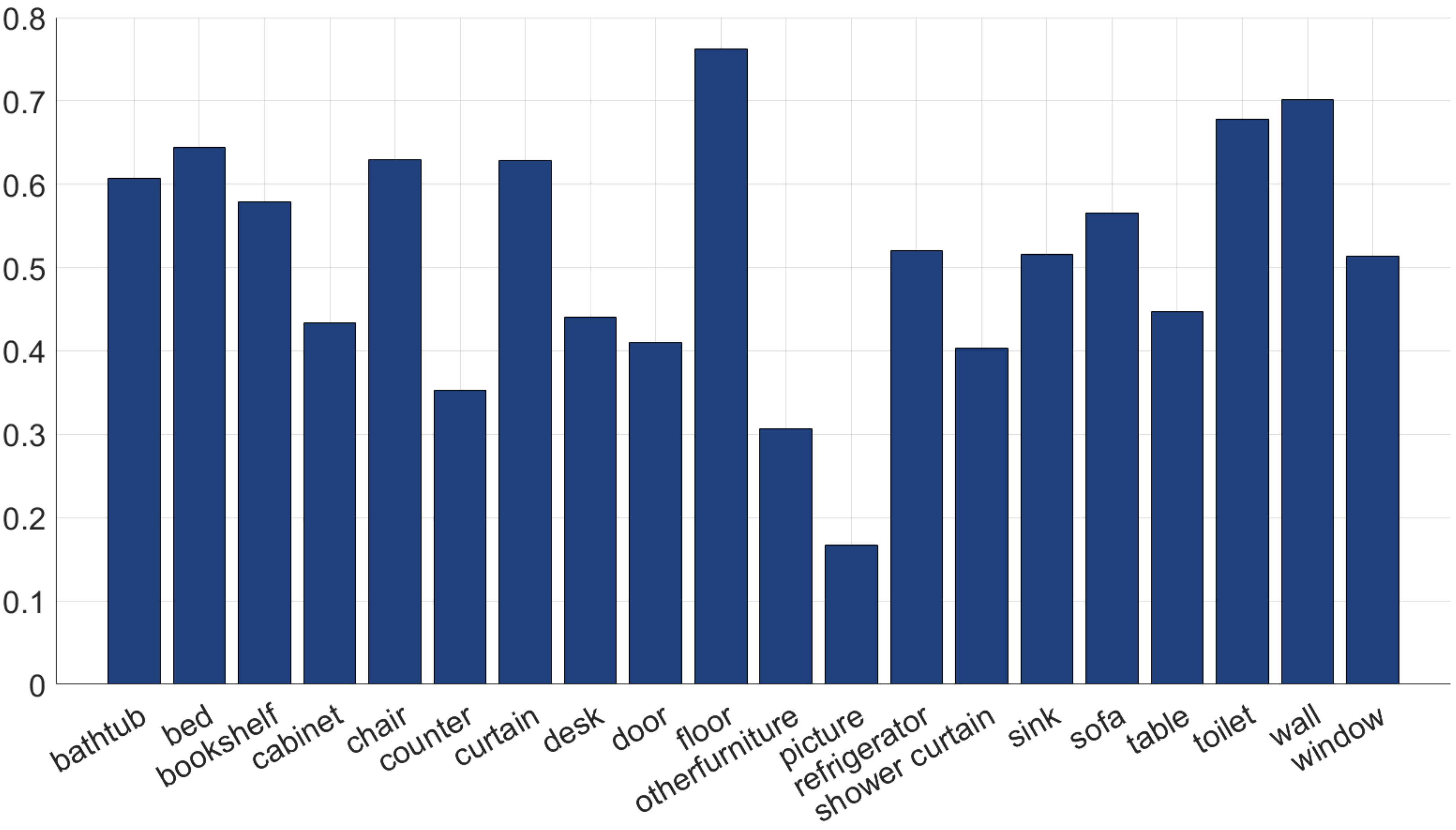}
    \caption{ScanNet per-class IoU (mIoU $=$ 0.515)}
    \end{subfigure}%
    \hspace{0.2cm}
    \begin{subfigure}{0.24\textwidth}
    \includegraphics[width=\linewidth]{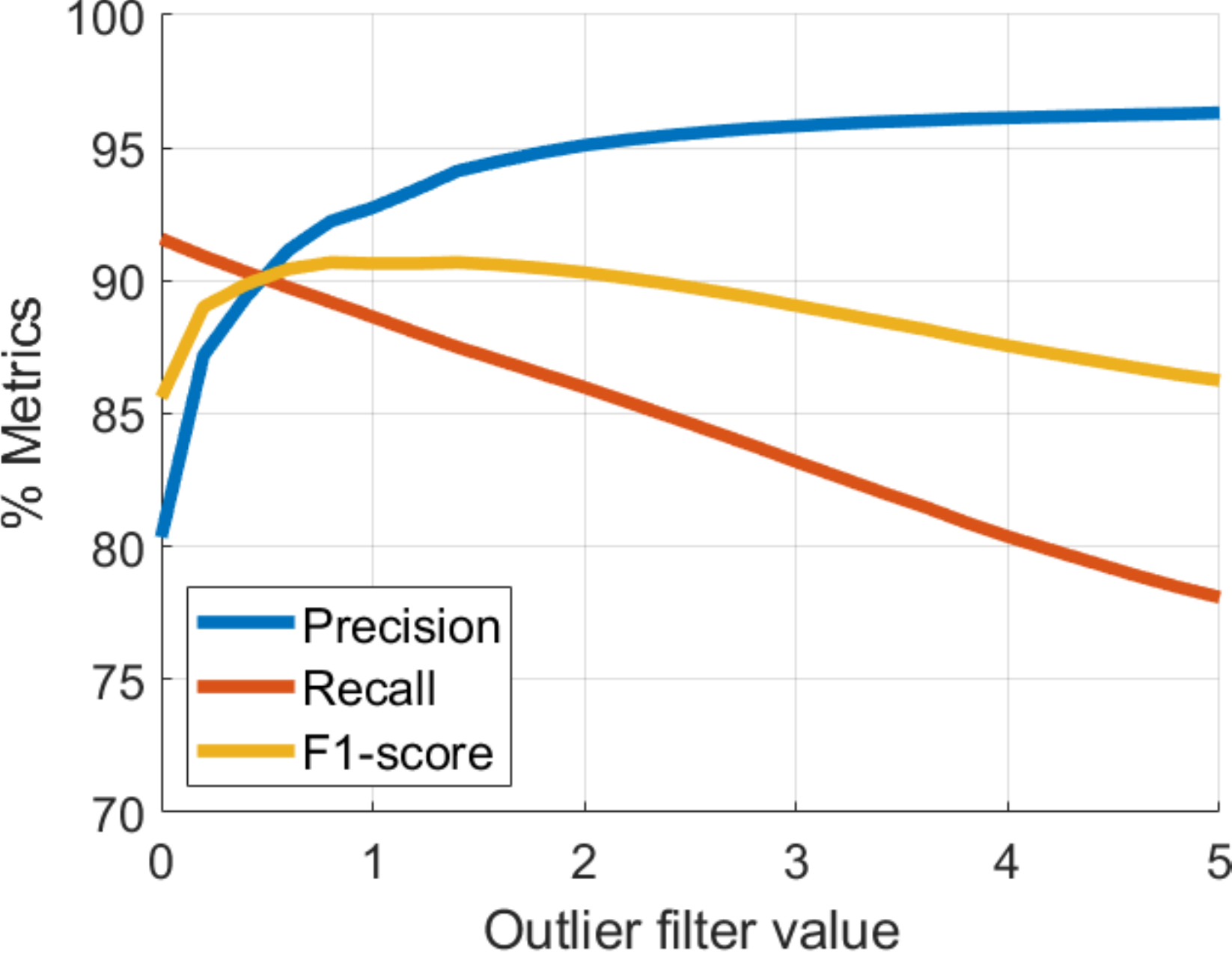}
    \caption{Outlier filter value}\label{fig:out_filter_val}
    \end{subfigure}%
    \hspace{0.3cm}
    \begin{subfigure}{0.32\textwidth}
    \includegraphics[width=\linewidth]{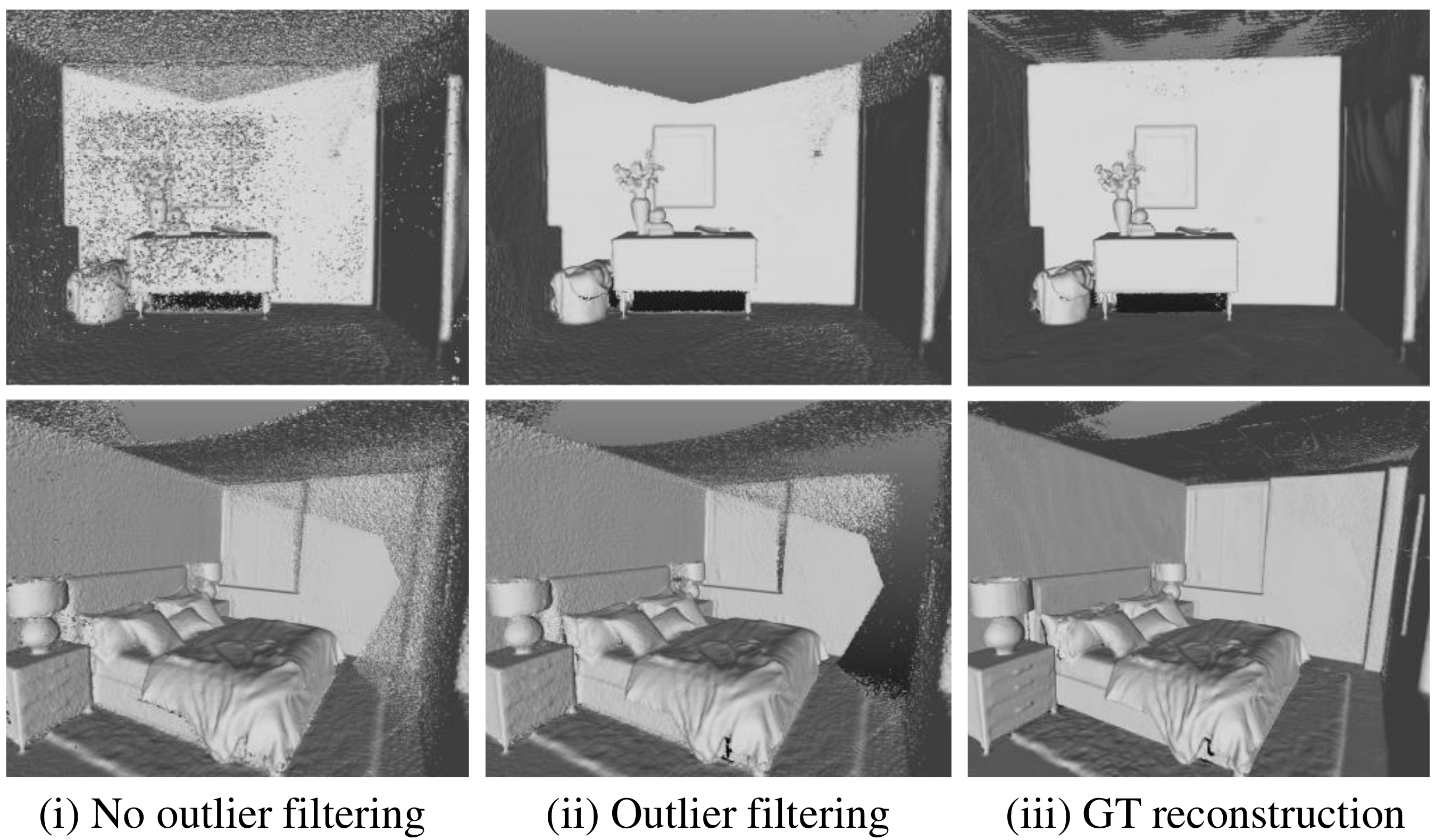}
    \caption{Qualitative Results of Outlier Removal}\label{fig:mesh_comp}
    \end{subfigure}
    \caption{(a) ScanNet per-class IoU. (b) Effect of outlier filter value on the different accuracy metric used for experiments. While the precision increases with the filter threshold, the other metric decrease. c) Effects of outlier filtering in the depth fusion pipeline for online 3D reconstruction: (i) without outlier filtering, (ii) with outlier filtering value 2, (iii) using ground truth depth maps.}
    \label{fig:extra_exp}
\end{figure*}

\formattedparagraph{(b) 3D Semantic Segmentation Evaluation.}
We evaluated our 3D semantic segmentation results on the ScanNet dataset. Our approach scored a mean IoU of 0.515. To our knowledge, our approach is among the early online deep-learning framework to jointly solve 3D semantic segmentation and scene reconstruction. Fig. \ref{fig:extra_exp}(a) shows per-class IoU scores on ScanNet. For comparative results, refer to the leaderboard\footnote{{\normalfont\url{http://kaldir.vc.in.tum.de/scannet\_benchmark/result\_details?id=766}}}.

Fig. \ref{fig:seg_comp} (top row) shows the per-class IoU on Replica test scene. To show our IoU results on the scene sequence, we synthesized a possible upper bound on the IoU achievable by projecting the ground truth 2D semantics to 3D. The light-blue color shows the upper bound, and the dark blue shows the IoU using our method over different classes. Fig. \ref{fig:seg_comp} (bottom row) show the visual results for the same.

\formattedparagraph{(c) Ablation Study.}

\noindent
\textit{(i) Performance with the change in number of network's parameter:}
Concatenating the semantic frame to the network inputs and removing the routing step led to Fusion Network v1 (Table \ref{tab:fusion_design}.C), which improves the mean F1-score by 2.1\% over RoutedFusion (Table \ref{tab:fusion_design}.A). By designing a more powerful encoder to exploit the semantics better and deal with the input noise, we proposed Fusion Network v2 (Table \ref{tab:fusion_design}.F), which further improves the F1-score by 4.6\%. Finally, by processing the semantic and depth frames with two separate encoders, we designed Fusion Network v3 (Table \ref{tab:fusion_design}.G), which further improves the F-score by 1.3\%. Table \ref{tab:fusion_design} shows the variation in 3D reconstruction quality with the number of parameters in the network design.

\begin{table}[h]
    \centering%
    \resizebox{\columnwidth}{!}
    {
    \begin{threeparttable}
    \begin{tabular}{c|c|c|c|c|c|c|c|c|c|c}
    \hline
        \multicolumn{2}{c|}{} & \multicolumn{3}{c|}{room\_1} &  \multicolumn{3}{c|}{office\_0} & \multicolumn{3}{c}{office\_2} \\
        \hline
        & Params & Prec. & Recall & F1 & Prec. & Recall & F1 & Prec. & Recall & F1 \\
        \hline
        A & 80k & 85.73 & 79.52 & 82.51 & 86.95 & 81.66 & 84.22 & 81.94 & 81.52 & 81.73 \\
        B & 88k & 87.55 & 75.71 & 82.87 & 89.73 & 84.73 & 85.54 & 84.95 & 81.63 & 83.26 \\ 
        C & 80k & 90.74 & 77.42 & 83.55 & 91.94 & 81.85 & 86.60 & 84.44 & 82.49 & 83.45 \\
        D & 264k & 94.80 & 80.23 & 86.91 & 93.61 & 83.80 & 88.44 & 89.63 & 87.76 & 88.69 \\
        E & 404k & 96.43 & 80.03 & 87.47 & 97.44 & 83.49 & 89.93 & 88.33 & 88.37 & 88.35 \\
        F & 399k & 98.05 & 79.46 & 87.78 & 98.22 & 83.50 & 90.26 & 91.16 & 87.45 & 89.27 \\
        G & 572k & \textbf{98.35} & \textbf{80.23} & \textbf{88.37} & \textbf{98.86} & \textbf{84.54} & \textbf{91.14} & \textbf{94.76} & \textbf{88.54} & \textbf{91.55} \\
    \hline
    \end{tabular}
    \smallskip
    \caption{(A) RoutedFusion, (B) GT semantics concatenated to the input, (C) no routing step (Fusion Network v1), (D) deeper encoder and 1 eASPP block \cite{valada2019self}, (E) 2 eASPP blocks \cite{valada2019self}, (F) 2 efficient vortex pooling blocks (Fusion Network v2), (G) different encoders for semantic and depth frames (Fusion Network v3). Note: Efficient Atrous Spatial Pyramid Pooling (eASPP) architecture was proposed in \cite{valada2019self}.}\label{tab:fusion_design}
    \end{threeparttable}
    }
\end{table}

\noindent
\textit{(ii) Dependency on semantics:} 
To understand the benefit of 2D semantic prior on overall 3D reconstruction accuracy, we noted our network's performance under three different settings using couple of scenes: (A) Using the ground-truth 2D semantic label (B) Using 2D semantic label predicted using our segmentation network. (C) no 2D semantic segmentation prior is used. Table \ref{tab:semantic_comparison} shows the statistical results for this experiment. Clearly, (A) performs the best; but, the difference between (A) and (B) is very less. When no semantic is used (C) the performance degrades showing the benefit of using semantics to improve online 3D reconstruction.

\vspace{-0.3cm}
\begin{table}[h]
    \centering
    \resizebox{0.85\columnwidth}{!}
    {
    \begin{threeparttable}
    \begin{tabular}{c|c|c|c|c|c|c|c}
    \hline
        \multicolumn{2}{c|}{} & \multicolumn{3}{c|}{room\_1} & \multicolumn{3}{c}{office\_2} \\
        \hline
        \footnotesize
        & Params & Prec. & Recall & F1  & Prec. & Recall & F1 \\
        \hline
        A & 572k & \textbf{98.35} & \textbf{80.23} & \textbf{88.37} & \textbf{94.76} & \textbf{88.54} & \textbf{91.55} \\
        B & 572k & \textbf{\textcolor{brown}{97.22}} & \textbf{\textcolor{brown}{79.99}} & \textbf{\textcolor{brown}{87.72}} & \textbf{\textcolor{brown}{92.35}} & \textbf{\textcolor{brown}{87.48}} & \textbf{\textcolor{brown}{89.85}} \\
        C & 360k & 97.00 & 78.54 & 86.80 & 88.39 & 86.35 & 87.36  \\
        \hline
    \end{tabular}
    \end{threeparttable}
    }
    \smallskip
    \caption{Fusion Network v3 trained with: (A) ground truth semantics, (B) predicted semantics, (C) no semantics.}%
    \label{tab:semantic_comparison}
\end{table}
\vspace{-0.3cm}
\noindent
\textit{(iii) Effect of outlier filtering value:} Outlier filtering is vital for high-quality depth fusion. Yet, unreasonable outlier filtering value can lead to the loss of high-frequency surface geometry. Fig.\ref{fig:extra_exp}(b) shows the effect of outlier filter value on the precision, recall, and F-score of the reconstructed scene. Fig. \ref{fig:extra_exp}(c) provide the qualitative comparison 3D reconstruction results obtained before and after outlier filtering.

\section{Conclusion}
\noindent
This paper proposed an online approach that enables joint inference of 3D structure and 3D semantic labels of an indoor scene in real-time. We showed that our approach provides better 3D reconstruction accuracy and inference time than the state-of-the-art online depth fusion approach \cite{Weder2020RoutedFusionLR}. Moreover, we demonstrated how to leverage the 2D semantic prior via a fusion network to further enhance 3D reconstruction accuracy. Subsequently, to seamlessly predict the 3D semantic label with better 3D reconstruction, we propose an update scheme module that improves the global semantic label volume representation by utilizing label prediction confidence and the local geometry update information. All in all, our approach performance and its flexibility to use on a machine with different computational resources show that it can be helpful for other important robot vision tasks.


%





\ifCLASSOPTIONcaptionsoff
  \newpage
\fi



%


\balance
\bibliographystyle{IEEEtran}
\bibliography{camera_ready}
\nocite{kumar2019dense}
\nocite{kumar2019superpixel}

%








\end{document}